\documentclass[conference]{IEEEtran}
\IEEEoverridecommandlockouts

\usepackage{cite}
\usepackage{amsmath,amssymb,amsfonts}
\usepackage{algorithmic}
\usepackage{graphicx}
\usepackage{textcomp}
\usepackage{xcolor}
\usepackage{hyperref}
\usepackage{threeparttable}
\usepackage{multirow}
\usepackage{colortbl}
\usepackage{pifont}
\usepackage{bm}
\usepackage{soul}
\usepackage{bbding}
\usepackage{wasysym}
\usepackage{booktabs}

\makeatletter
\newcommand{\linebreakand}{%
  \end{@IEEEauthorhalign}
  \hfill\mbox{}\par
  \mbox{}\hfill\begin{@IEEEauthorhalign}
}
\makeatother

\hypersetup{
  colorlinks=true,
  linkcolor=black,
  urlcolor=black,
  citecolor=black
}

\definecolor{mygray}{gray}{.94}

\def\BibTeX{{\rm B\kern-.05em{\sc i\kern-.025em b}\kern-.08emT\kern-.1667em\lower.7ex\hbox{E}\kern-.125emX}}

\begin{document}

\title{BitStopper: An Efficient Transformer Attention Accelerator via Stage-fusion and Early Termination

\thanks{Corresponding authors: Yang Hu (hu\_yang@tsinghua.edu.cn)}
}

\author{
\IEEEauthorblockN{Huizheng Wang}
\IEEEauthorblockA{\textit{School of Integrated Circuits} \\
\textit{Tsinghua University}\\
Beijing, China \\
wanghz22@mails.tsinghua.edu.cn}
\and
\IEEEauthorblockN{Hongbin Wang}
\IEEEauthorblockA{\textit{School of Integrated Circuits} \\
\textit{Tsinghua University}\\
Beijing, China \\
wanghb24@mails.tsinghua.edu.cn}
\and
\IEEEauthorblockN{Shaojun Wei}
\IEEEauthorblockA{\textit{School of Integrated Circuits} \\
\textit{Tsinghua University}\\
Beijing, China \\
wsj@tsinghua.edu.cn}
\linebreakand
\IEEEauthorblockN{Yang Hu}
\IEEEauthorblockA{\textit{School of Integrated Circuits} \\
\textit{Tsinghua University}\\
Beijing, China \\
hu\_yang@tsinghua.edu.cn}
\and
\IEEEauthorblockN{Shouyi Yin}
\IEEEauthorblockA{\textit{School of Integrated Circuits} \\
\textit{Tsinghua University}\\
Beijing, China \\
\textit{Shanghai Artificial Intelligence Laboratory}\\
Shanghai, China \\
yinsy@tsinghua.edu.cn}
}

\maketitle

\begin{abstract}
Attention-based large language models (LLMs) have transformed modern AI applications, but the quadratic cost of self-attention imposes significant compute and memory overhead. Dynamic sparsity (DS) attention mitigates this, yet its hardware efficiency is limited by the added prediction stage and the heavy memory traffic it entails. To address these limitations, this paper proposes BitStopper, a fine-grained algorithm-architecture co-design that operates without a sparsity predictor. First, a bit-serial enable stage fusion (BESF) mechanism is proposed to reuse and minimize the memory access by progressively terminating trivial tokens and merging the prediction stage into the execution stage. Second, a lightweight and adaptive token selection (LATS) strategy is developed to work in concert with the bit-level sparsity speculation. Third, a bit-level asynchronous processing (BAP) strategy is employed to improve compute utilization during the on-demand bit-grained memory fetching. Finally, an elaborate architecture is designed to translate the theoretical complexity reduction into practical performance improvement. Extensive evaluations demonstrate that, compared to state-of-the-art (SOTA) Transformer accelerators, BitStopper achieves $2.03\times$ and $1.89\times$ speedups over Sanger and SOFA, respectively, while delivering $2.4\times$ and $2.1\times$ improvements in energy efficiency.
\end{abstract}


\section{Introduction}\label{sec:introduction}
Large language models (LLMs) now underpin a wide range of applications, from automated content generation \cite{wei2022emergent,yuan2022wordcraft} to conversational assistants \cite{thoppilan2022lamda,zhang2024benchmarking}. With the rapid advances of GPT-series models \cite{radford2019language,kalyan2024survey,ye2023comprehensive}, improving the inference efficiency of LLMs has become a current research priority.

LLMs derive much of their effectiveness from the self-attention mechanism \cite{vaswani2017attention}, which captures global dependencies throughout the entire sequence. In each self-attention module, the Query (Q), Key (K), and Value (V) matrices serve to encode the pairwise dependencies between tokens. Despite its strong modeling capabilities, the attention mechanism remains computationally inefficient: its quadratic scaling with sequence length and the predominance of low-intensity element-wise operations impose substantial overhead. As a result, the attention has been widely demonstrated as the principal bottleneck to efficient LLM inference \cite{wang2025mcbp,kim2023full,fuad2023survey,zhao2023survey,zhou2024survey}.


\begin{figure}[t]
\centering
\includegraphics[width=\linewidth]{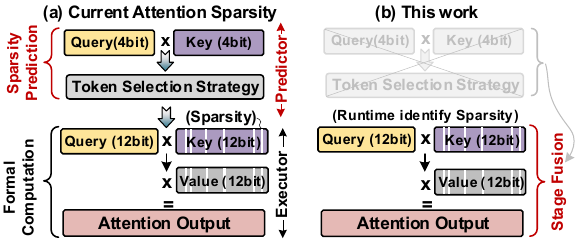}\vspace{-3mm}
\caption{Workflow comparison of (a) traditional DS works and (b) this work, where runtime identifies sparsity directly during \emph{formal computation} without an extra sparsity predictor.}
\label{fig:Predictor}\vspace{-5mm}
\end{figure}

To mitigate the inefficiencies inherent in attention, \emph{dynamic sparsity (DS) attention} \cite{ham20203,ham2021elsa,zhou2022energon,qu2022dota,qin2023fact,wang2024sofa,lu2021sanger,li2022accelerating,wang2021spatten,liu2022dynamic,yang2022dtatrans,fan2022adaptable} has emerged as a widely adopted approach. As shown in Fig. \ref{fig:Predictor} (a), the DS generally incorporates an initial \emph{prediction stage} in which a low-bitwidth predictor approximates the attention matrix to identify and eliminate low-correlation Q-K pairs. Afterward, the remaining Q-K pairs are processed by an executor operating at higher precision, which we refer to as \textit{formal computation stage}. 

However, upon revisiting current DS approaches, we observe that they largely target two issues: 1) Reducing the prediction workload; 2) Streamlining the identification of essential tokens. For example, methods such as Energon \cite{zhou2022energon}, FACT \cite{qin2023fact}, ELSA \cite{ham2021elsa} and SOFA \cite{wang2024sofa} aim to lower prediction cost through schemes such as iterative score refinement, log-domain shifting, binary hashing and distributed sorting. In parallel, designs like DOTA \cite{qu2022dota}, Sanger \cite{lu2021sanger} and LeOPArd \cite{li2022accelerating} simplify the identification of influential tokens by relying on fixed, empirically derived or heuristic threshold values.

\textbf{Unfortunately, our analysis shows that the additional sparsity predictor is what largely offsets the advantages of DS, an issue overlooked in prior works.} For instance, although SOFA \cite{wang2024sofa} successfully mitigates the computation overhead via logarithmic-domain processing, it nevertheless must fetch the entire set of Key vectors, causing the prediction stage to dominate the total power consumption by exceeding $75\%$. Although TokenPicker \cite{park2024token} and SpAtten \cite{wang2021spatten} attempt to address the prediction-stage I/O burden by employing progressive quantization, their relatively coarse segmentation granularity limits the achievable benefits. Likewise, SpARC \cite{cho2024sparc} and CLAT \cite{lee2025clat} partially mitigate this burden via offline clustering, yet the substantial calibration overhead limits their practical deployment. \emph{Overall, current DS approaches remain constrained by an external sparsity predictor whose heavy memory-access demands fundamentally restrict efficiency.}


\textbf{Our insights}: The root cause of the unjustifiable prediction cost stems from the severe decoupling between the sparsity predictor and the attention executor. This separation prevents the computational and memory access efforts expended in the predictor from being leveraged by the subsequent executor. 

\begin{figure}[t]
\centering
\includegraphics[width=\linewidth]{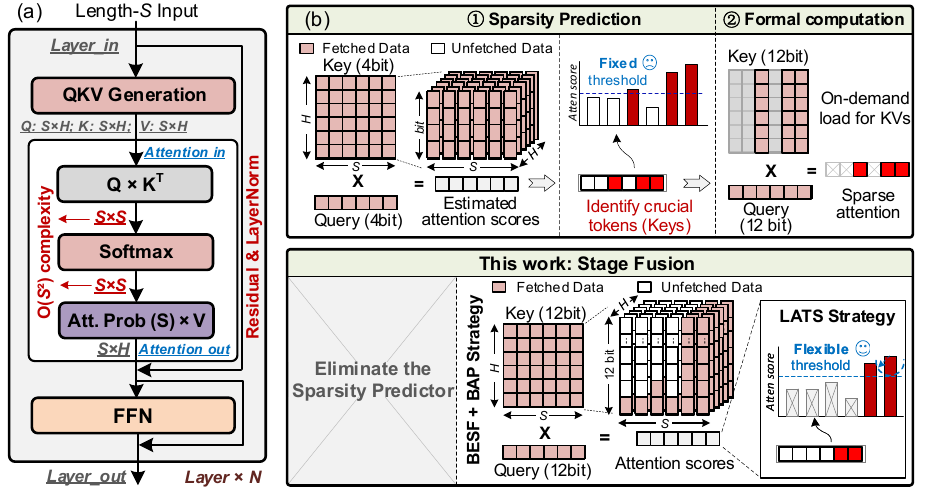}\vspace{-3mm}
\caption{(a) Architecture of Transformer-based LLMs. (b) The workflow of the current DS works. (c) Illustration of the BitStopper, featuring stage-fusion.}
\label{fig:Transformer}\vspace{-4mm}
\end{figure}

Inspired by bit-serial computing \cite{lee2018unpu,kam2025panacea,im2023sibia,guo2025transitive}, we propose a bit-serial enabled stage fusion (BESF) strategy to eliminate the extra sparsity predictor. The idea is to interleave prediction with execution and reduce memory traffic by incrementally processing Key vectors at the bit-plane level. Specifically, BESF begins with the most significant bit of each Key vector to estimate  $\mathbf{Q}$$\times$$\mathbf{K}^T$ correlations. Once a token (i.e., a vector of Key) is judged unimportant, the system immediately halts further bit-plane loading for that token. As a result, the executor only completes the remaining computation for the surviving Q-K pairs and produces the final output by accumulating their partial results.


\textbf{Despite its promise, designing a BESF-style DS accelerator is far from trivial. It faces two key challenges:} 1) The absence of adaptive and accurate token selection policies suited to bit-granular scores. 2) On-demand and fine-grained processing exposes memory-access latency frequently, which may degrade hardware utilization. 

To address the above challenges, this paper introduces BitStopper, a novel software-hardware co-design to accelerate attention at the bit granularity. Our contributions includes:

\textbf{1) Bit-serial enabled Stage Fusion (BESF) Strategy to eliminate the need for an additional prediction stage.} Leveraging the bit-serial computing, we dedicate the BESF strategy, which seamlessly fuses the prediction stage into the formal computation stage, removing the prediction overhead. 

\textbf{2) Lightweight Adaptive Token Selection
(LATS) strategy for bit-grained speculation while ensuring pruning accuracy}. We first introduce the \emph{bit-level uncertainty margin}, which enables the assessment of inner product fluctuation ranges. Leveraging this, the LATS is designed to identify trivial tokens in an adaptive, progressive, bit-grained manner.

\textbf{3) Bit-level asynchronous processing (BAP) to enhance hardware utilization}. To address resource underutilization from fine-grained bit-level computation, we propose an efficient asynchronous execution scheme that breaks the strict sequential dependency of traditional bit-serial processing, effectively hiding the DRAM access latency.

\textbf{4) Custom accelerator design for practical acceleration}. To translate the theoretical improvement into practical performance gains, we design a customized accelerator, BitStopper, to support these proposed strategies. BitStopper achieves $2.4\times$ and $2.1\times$ improvements in energy efficiency over state-of-the-art (SOTA) DS accelerators, Sanger and SOFA, respectively.

\begin{figure}[t]
\centering
\includegraphics[width=\linewidth]{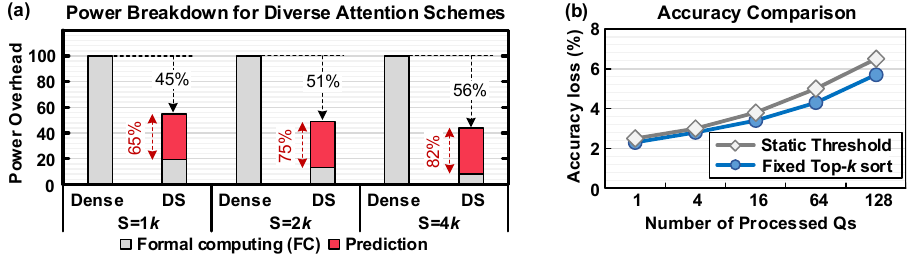}\vspace{-3mm}
\caption{(a) Comparison of power distribution between dense attention and DS attention on TSMC 28 nm. (b) Accuracy of various token-selection strategies.}
\label{fig:Motivation}\vspace{-5mm}
\end{figure}

\section{Background and Motivation}\label{sec:Motivation}
\subsection{Transformer Architecture}
Fig. \ref{fig:Transformer} (a) presents the standard Transformer structure, composed of three stages: QKV generation, self-attention and feed-forward network (FFN). First, a length-$S$ sequence embedded and projected into the Query (Q), Key (K), Value (V) spaces. The self-attention mechanism then computes contextual dependencies through the operation in Eq. \eqref{eq:self_attention}. The resulting attention outputs are subsequently processed by an FFN with two fully connected layers to produce the final representation.
\begin{equation}
\mathbf{O} = {\rm softmax} \left( \mathbf{A}/{\sqrt{d_{\rm h}}}\right)\times \mathbf{V}, ~{\rm where}~ \mathbf{A}=\mathbf{QK}^T.
\label{eq:self_attention}   
\end{equation}

\subsection{Dynamic Sparsity in Self-Attention}\label{subsec:Attention_Sparsity}
Although self-attention allows a model to examine interactions across all tokens in a sequence, only a subset of these interactions carries meaningful contextual information \cite{song2024tsacc,zhao2024hardware,wang2023cta,hong2022dfx,park2024token}. Common function words, for example “a” or “the”, often contribute negligibly to contextual reasoning and consequently receive extremely small softmax weights in Eq.~\eqref{eq:self_attention}. This imbalance in token significance has motivated a line of DS work that attempts to exploit attention sparsity at runtime. As shown in Fig.~\ref{fig:Transformer}(b), DS works typically follow a two-step workflow. During the \textit{sparsity prediction stage}, a low-precision approximation of $\mathbf{QK}^{T}$ (e.g., 4-bit) is computed to identify informative Q-K pairs via top-$k$ ranking or thresholding. During the subsequent \textit{formal computation stage}, the final attention computation is carried out at higher precision (e.g., 12-bit) on only the retained Q-K pairs.


\subsection{Motivation}\label{subsec:Motivation}
However, upon re-examining existing DS methods, we identify two significant limitations:

\begin{figure}[t]
\centering
\includegraphics[width=0.95\linewidth]{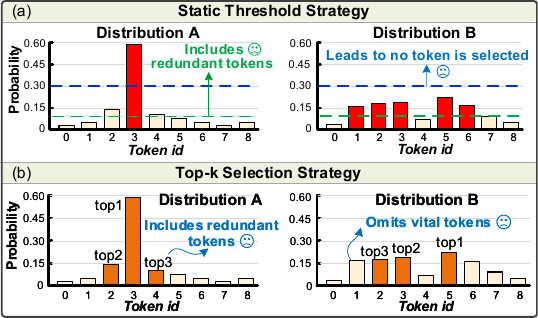}\vspace{-3mm}
\caption{Fundamental limitations of current token selection strategies.}
\label{fig:Distribution}\vspace{-6mm}
\end{figure}

\emph{\textbf{1) Prohibitive Sparsity Prediction Overhead: }} 
As revealed by Fig.~\ref{fig:Motivation} (a), while DS lowers the overall power usage by exploiting token sparsity relative to dense attention, the cost incurred by its sparsity-prediction stage ultimately emerges as the primary source of total power consumption. For a $2k$-length input, the prediction stage, which performs the approximate evaluation of $\mathbf{QK}^T$ and identifies important tokens, draws nearly $3\times$ more power than the formal computation stage. This excessive cost largely stems from the need to access and process the entire Key matrix (size $S$$\times$$H$). Moreover, this burden remains irreducible by sparsity, as the Key tensor must be fully accessed. When the sequence length increases to $4k$, the power ratio of prediction escalates to a staggering $4.7\times$.

\emph{\textbf{2) Inflexible Token Selection Strategies:}} Fig. \ref{fig:Distribution} illustrates the inflexibility of current token selection strategies using two different attention distributions. For the predefined threshold comparison in Fig.~\ref{fig:Distribution} (a), a high threshold (blue line) effectively identifies the critical token (token 3) in Distribution A (Dist A) but fails to select any vital tokens in Dist B. In contrast, a lower threshold (green) captures vital tokens in Dist B, but also retains irrelevant tokens (i.e., 2, 4) in Dist A. Additionally, Fig. \ref{fig:Distribution} (b) depicts the limitations of current top-$k$ selection. For example, with top-$3$ selection, unnecessary tokens (i.e., 2, 4) are retained in Dist A, whereas important tokens in Dist B, such as tokens $1$ and $6$, are mistakenly filtered out. Further, Fig. \ref{fig:Motivation} (b) depicts how the accuracy of both approaches declines as the number of Qs increases. It is evident that both strategies suffer from evident accuracy loss due to their static nature, which fails to adapt to the diverse distribution of attention scores ($\mathbf{Q}_i\mathbf{K}^T$) across different Qs.

\section{Algorithm Optimizations of BitStopper}\label{sec:BUPC}
To address the issues of high prediction overhead and rigid token-selection behavior, we propose three key techniques. First, the BESF mechanism fuses the prediction stage into execution stage to eliminate redundant computation and reduce memory traffic. Second, the lightweight and adaptive LATS scheme enhances the identification of critical tokens across diverse distributions. Finally, the BAP strategy improves the hardware efficiency of bit-grained processing.

\subsection{Bit-Serial Enabled Stage Fusion
(BESF) Mechanism }\label{subsec:Eager_Processive_Estiamtion}
The BESF builds on the observation that attention scores often exhibit large disparities, meaning that weak QK interactions can be dismissed using only a small portion of the bit information. This allows high-precision processing to be reserved exclusively for genuinely influential Q-K pairs. 


Leveraging this property, BESF performs bit-incremental pruning: it derives coarse attention estimates from high-order bit planes and progressively refines the candidate set as lower-order bits are revealed. This incremental narrowing enables early termination of redundant IO and computation cost. Meanwhile, the partial results produced in low-bit prediction can be reused during the formal computation stage, thereby eliminating the standalone prediction overhead.

\begin{figure}[t]
\centering
\includegraphics[width=\linewidth]{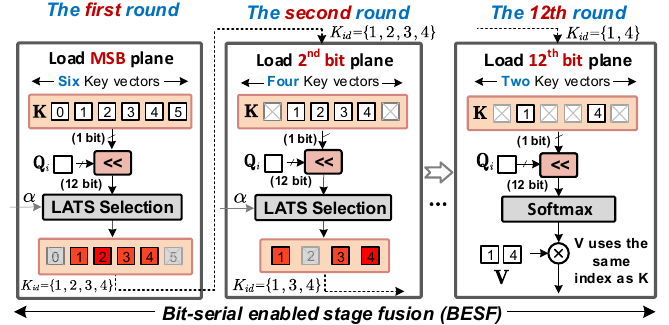}\vspace{-3mm}
\caption{Illustration of the bit-serial enabled stage fusion (BESF) mechanism.}
\label{fig:BSF}\vspace{-5mm}
\end{figure}

Fig. \ref{fig:BSF} outlines the overall procedure. 1) For each query $\mathbf{Q}_i$ (with 12-bit quantization), BESF begins by computing approximate dot-product scores using only the MSB of the Keys. Keys (e.g., $\{1,2,3,4\}$) whose potential max scores exceed an initial threshold are retained for further distingish. Subsequent rounds evaluate additional bit planes, repeatedly filtering the retained set based on the updated partial scores. In the illustrated example, the retained set $\{1,2,3,4\}$ is tightened to $\{1,3,4\}$ and eventually to surviving Keys $\{1,4\}$ after multiple refinement rounds. Once the sparse $\mathbf{S}$ is generated, full-precision $\mathbf{S}$$\times$$\mathbf{V}$ computation (12-bit) is performed using the Vs corresponding to the indices of the selected final Keys.   


\subsection{Lightweight Adaptive
Token Selection (LATS) Strategy}
However, it remains difficult to pinpoint potentially important Keys at each bit round in a lightweight manner that can still adapt to the shifting distribution of attention scores.


To this end, we propose the LATS strategy. We begin by analyzing the mathematical behavior of the softmax function and show that it naturally supports lightweight max-oriented decision processes. Consider a two-element vector $[a_0,a_1]$ passed through the softmax, where $a_1$ is the larger value and can be expressed as $a_1$$=$$a_0$$+$$\delta$. From Eq.~\eqref{eq:softmax}, the probability assigned to $a_0$ diminishes exponentially as the gap $\delta$ increases. This reveals a fundamental property: \emph{within any input vector, an element’s softmax probability is tightly linked to its distance from the maximum}. A larger deviation from the maximal input results in a proportionally smaller impact on the output distribution. 
\begin{equation}
\!\!\mathrm{softmax}(a_0) \!=\! \frac{e^{a_0}}{e^{a_0} \!+\! e^{a_0 + \delta}}\!=\!\frac{1}{1 \!+\! e^{\delta}}\!=e^{-\delta}\! \cdot\! \frac{1}{1 + e^{-\delta}}\!<\!e^{-\delta}
\label{eq:softmax}
\end{equation}

Building on this observation, we design an adaptive thresholding scheme that derives its decision boundary from the max attention score at each bit prediction round. Its formulation is given in Eq. \eqref{eq:DFD}, where $A_{i,:}^r$ denotes the attention score at the $r$-th prediction round and $\eta_i$ is the threshold for the $i$-th Query. In each round, the attention scores are compared against $\eta_i$ and only the Keys with scores exceeding this threshold are preserved. Based on extensive experiments, we set the default \emph{radius} to 5. To accommodate diverse accuracy–efficiency trade-offs, we further incorporate a hyperparameter $\alpha\in[0, 1]$ which scales the threshold and thereby controls the pruning aggressiveness of each round. Because the pruning threshold $\eta_i$ is directly inferred from the underlying attention distribution, the method naturally adapts to score variations and achieves high accuracy in selecting vital tokens.
\begin{equation}
\eta _{i}\!=\!\max \left(A_{i,:}^{r} \right) - \!\alpha \times radius, ~\alpha\in [0,1].
\label{eq:DFD}
\end{equation}


\begin{figure}[t]
\centering
\includegraphics[width=\linewidth]{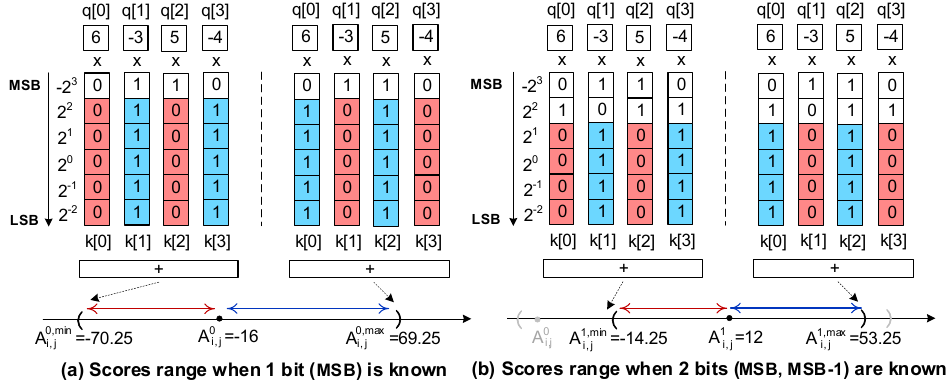}\vspace{-3mm}
\caption{Illustration for bit margin of dot-product $\mathbf{Q}_i\times \mathbf{K}_j^T$ ($\mathbf{A}_{i,j}$).}
\label{fig:bit_prediction}\vspace{-4mm}
\end{figure}

However, a subsequent challenge lies in how to accurately estimate the possible values of the dot product $\mathbf{Q}_i\mathbf{K}_j^T$ (i.e., $A_{i,j}$), with only partial bits of the Key vectors. To this end, we introduce a \emph{bit-level uncertainty margin}, which quantifies the potential variation introduced by the remaining (unprocessed) bits. Specifically, for an $N$-bit integer $c_{N-1}c_{N-2}...c_0$ with 2's complement number format, its value $x$ is:
\begin{equation}
x=-c_{N-1}2^{N-1}+\sum\nolimits_{i=0}^{N-2}c_i2^i.
\end{equation} 

In this format, all bits except the sign bit ($c_{N-1}$) contribute a non-negative value, meaning that each additional bit can only increase or maintain the magnitude of the number. Based on this,  Fig.~\ref{fig:bit_prediction} (a)(b) examplifies the bit margin. In this example, $\mathbf{Q}_i$ retains all six bits, while $\mathbf{K}_j$ has a fraction of $6$ bits: 1\,bit in Fig.~\ref{fig:bit_prediction} (a) and 2\,bits in Fig.~\ref{fig:bit_prediction} (b). For elements of $\mathbf{Q}_i$ that are positive, setting the unknown bits of $\mathbf{K}_j$ to 1 (shown in blue), or to 0 (shown in red) if the element of $\mathbf{Q}_i$ is negative, yields a potential maximum score $A_{i,j}^{r,\max}$, as it accounts only for positive contributions. Conversely, flipping the unknown bits yields the potential minimum score $A_{i,j}^{r,\min}$.

\begin{figure}[t]
\centering
\includegraphics[width=\linewidth]{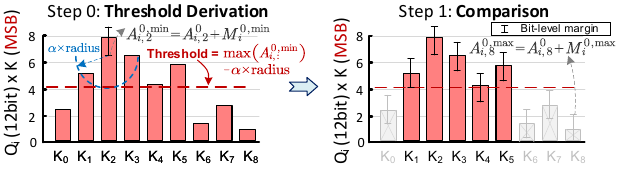}\vspace{-3mm}
\caption{Illustration of the bitwise LATS strategy.}
\label{fig:Illustration}\vspace{-6mm}
\end{figure}

Using the concept, we first utilize the low bound $A_{i,:}^{r,\min}$ to derive the $\eta_i$. Then, when making pruning decisions for a $\mathbf{K}_j$, its potential maximum score $A_{i,j}^{r,\max}$ at the $r$-th round is compared with the threshold $\eta_i$. As exemplified in Fig.\ref{fig:Illustration}, for speculation by the MSB of Keys (i.e., $r$$=$$0$), during \emph{Threshold Derivation}, the max (i.e., $A_{i,2}^{0,\min}$) among all lower bounds is selected, which is then subtracted by $\alpha$$\times$$radius$ to derive the threshold. In subsequent \emph{Comparison}, all upper bounds, i.e., $A_{i,j}^{0,\max}$, are compared with this threshold to decide on pruning.

\subsection{Bit-Level Asynchronous Processing (BAP) Strategy}\label{subsec:asynchronous}
The BESF workflow determines token pruning decisions from incrementally accumulated partial scores, which are computed as successive bit planes of each Key vector become available. For any token that remains unpruned, the system will retrieve the next bit plane from DRAM to further refine its score. However, when these off-chip memory fetches are issued and serviced strictly in sequence, the resulting exposed access latency can leave the compute units idle, ultimately degrading overall hardware utilization.



To address this issue, we propose a bit-level asynchronous processing (BAP) strategy. Here, \textit{asynchronous} indicates bit planes are no longer bound to arrive or be processed in a fixed order. Instead, computation proceeds whenever the required data becomes available.

\begin{figure}[t]
\centering
\includegraphics[width=\linewidth]{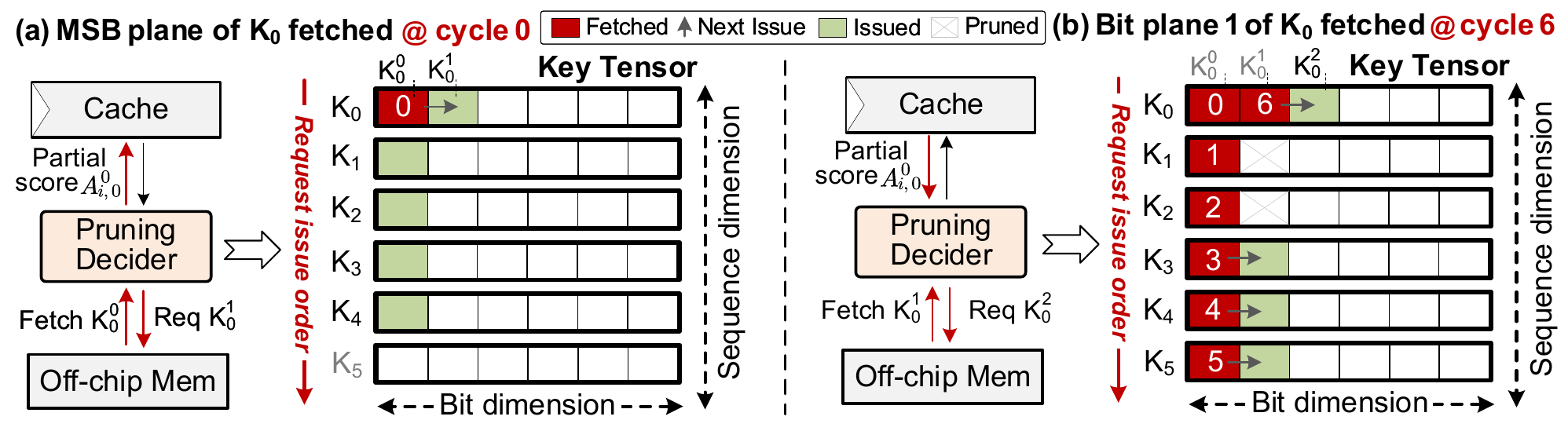}\vspace{-3mm}
\caption{Bit-level asynchronous processing (BAP) strategy.}
\label{fig:OOBE}\vspace{-6mm}
\end{figure}

As shown in Fig. \ref{fig:OOBE}, the BAP procedure begins by fetching only the most significant bit planes of the Key vectors, which provide the initial information needed for score estimation. As soon as one of these bit planes is returned from DRAM, its partial dot-product contribution is computed to determine whether the corresponding Key remains a candidate. If the Key survives this test, the following bit plane of the same Key is requested. Meanwhile, BAP stores intermediate results in a small on-chip buffer. Otherwise, BAP immediately moves on to the first bit plane of the next Key. When a later bit plane arrives (e.g., the loaded $K_0^1$ in Fig. \ref{fig:OOBE} (b)) arrives from DRAM, the previously buffered partial score ($A_{i,0}^0$) is retrieved and updated with the newly computed partial score, after which the BAP strategy repeats the same process. By continuously alternating between score refinement and new bit-plane requests, the compute units remain active even under long off-chip latency, substantially improving utilization during bit-level sparsity prediction.


\section{BitStopper Accelerator}\label{sec:Hardware}

\subsection{Overall Architecture}
\begin{figure*}[t]
\centering
\includegraphics[width=\linewidth]{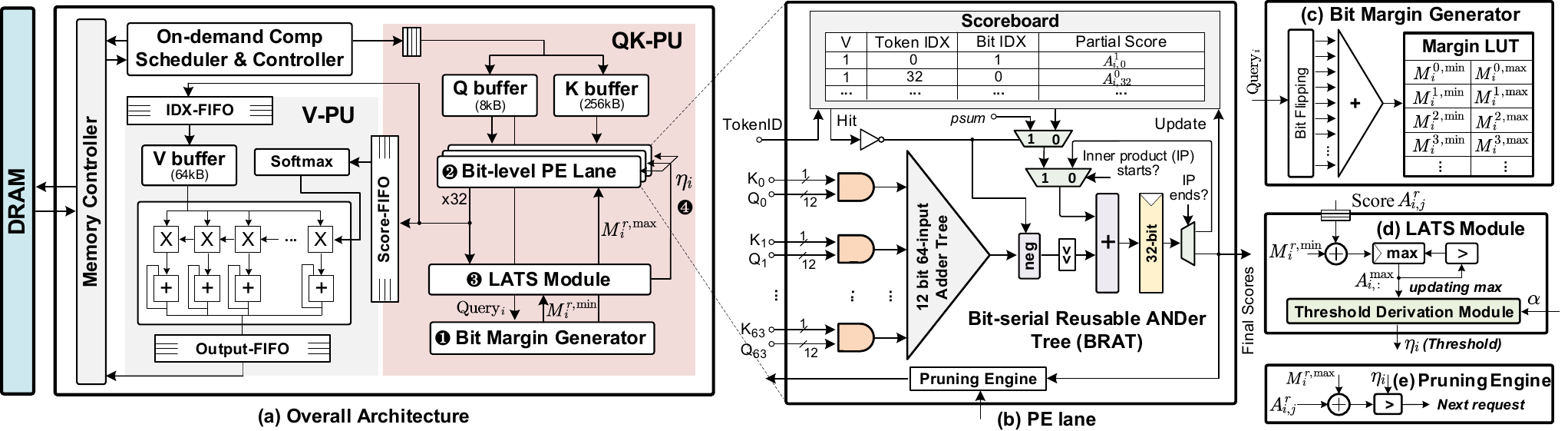}\vspace{-4mm}
\caption{(a) Overview of the BitStopper accelerator. (b) PE lane architecture. (c) Bit margin generator. (d) LATS module. (e) Pruning engine.}
\label{fig:Overall_architecture}\vspace{-6mm}
\end{figure*}

Fig. \ref{fig:Overall_architecture} (a) depicts the overall architecture of the BitStopper, which incorporates two major components:

\textbf{1) Query-Key Processing Unit (QK-PU)}: This unit performs the sparse $\mathbf{QK}^T$ computation directly, eliminating the need for an external sparsity predictor. It integrates $32$ bit-level PE lanes, a LATS Module and a Bit Margin Generator, which collaboratively support the BESF, LATS and BAP strategies.  

\textbf{2) Value Processing Unit (V-PU)}: The V-PU computes the final attention results from the Value vectors aligned with the Keys preserved by the QK-PU. It integrates a softmax module and a multiply-accumulate (MAC) array, which performs a weighted summation over the corresponding Value vectors. To ensure efficient pipelining, the MAC array is designed to perform $64$ INT12 MAC operations in a single cycle. 

\textbf{Overall Dataflow}: In BitStopper, self-attention is performed with a 12-bit per tensor quantization, where each Key vector is decomposed into twelve 1-bit planes. For a given Query $\mathbf{Q}_i$, the QK-PU first employs the BESF to obtain the sparse attention scores, with the aid of LATS and BAP strategies. Then, the V-PU generates the final attention outputs. As depicted in Fig.~\ref{fig:Overall_architecture} (a), its detailed process is as follows:

First (\ding{182}), before the $\mathbf{Q}_i\mathbf{K}^T$ computation, the Bit Margin Generator initializes twelve margin pairs $(M_{i}^{r,\min}, M_{i}^{r,\max})$, where $r$\,$\in$\,$[0,11]$, based on the input $\mathbf{Q}_i$. Each pair corresponds to a specific bit plane. These bit margin pairs are then stored in a lookup table (LUT), as shown in Fig. \ref{fig:Overall_architecture} (c). Following this (\ding{183}), $32$ PE lanes perform the dot product for $\mathbf{Q}_i\mathbf{K}^T$ and in an asynchronous manner in parallel. Then (\ding{184}), the LATS Module, as depicted in Fig. \ref{fig:Overall_architecture} (d), calculates the pruning threshold $\eta_i$ using the maximum value among the all current minimum score margins $A_i^{\min}$ with the LATS logic (Eq. \eqref{eq:DFD}). Finally (\ding{185}), the pruning threshold $\eta_i$ is broadcast to all PE lanes, enabling the evaluation of whether the score of the token $j$ satisfying $A_{i,j}^{r,\max}$$>$$\eta_i$, as depicted in Fig. \ref{fig:Overall_architecture} (e). If true, the PE lane requests the next bit plane for further computation. Otherwise, the token $j$ is immediately pruned. This process is repeated until the LSB is reached. The remaining scores are sent to the V-PU to produce final attention outputs.

\subsection{Microarchitecture of Bit-level PE Lane}
Fig. \ref{fig:Overall_architecture} (b) presents the microarchitecture of the bit-level PE lane. To support LATS and BAP execution, the PE lane incorporates two key modules in addition to a bit-serial reusable ANDer tree (BRAT): 1) The Pruning Engine evaluates intermediate results and determines both the pruning outcome and the next bit plane to be retrieved. 2) The Scoreboard stores the partial scores $S_{i,j}^r$ for tokens that remain unpruned, allowing subsequent rounds to reuse these values efficiently. 


\begin{figure}[t]
\centering
\includegraphics[width=\linewidth]{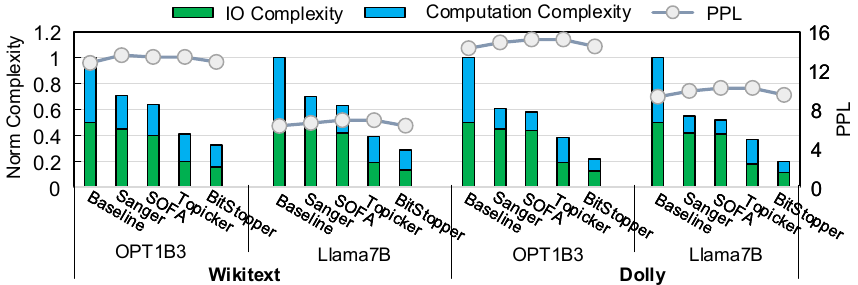}\vspace{-4mm}
\caption{Normalized complexity and PPL comparison of different DS works.}
\label{fig:memory_access_reduction}\vspace{-4mm}
\end{figure}

These modules collaborate to enable efficient early pruning. The process begins with the BRAT computing the partial dot product $\Delta A_{i,j}^r$ from a 12-bit vector $\mathbf{Q}_i$ and the 1-bit plane vector $\mathbf{K}_j^r$. In parallel, the Scoreboard is accessed with the token index $j$. If a partial score ($A_{i,j}^{r-1}$) from the prior bit plane is available, it is retrieved and added with the newly computed value to form the updated partial score $A_{i,j}^r$ (i.e., $A_{i,j}^{r-1}+\Delta A_{i,j}^r$). Otherwise, if no prior entry exists, this indicates that the current bit plane is the MSB. In this case, the partial score $\Delta A_{i,j}^r$ is directly written into Scoreboard, and the Hit signal is deasserted to show that no value can be reused.

Following this, the Pruning Engine performs the pruning operation. It receives the upper margin $M_{i}^{r,\max}$ along with the partial score $A_{i,j}^r$, and checks whether the condition $A_{i,j}^r+M^{r,\max}_i$$>$$\eta_i$ is satisfied. If the condition is met, it proceeds to fetch the next bit plane of $\mathbf{K}_j$, i.e., $\mathbf{K}_j^{r+1}$ and updates the partial score stored in the Scoreboard. Otherwise, it issues an eviction signal to remove the token’s entry from Scoreboard and then requests the next Key vector from DRAM.

As shown in Fig. \ref{fig:Overall_architecture} (d), the LATS module first retrieves the stored scores from the register and combines them with the min margin $M_i^{r,\min}$ to derive their lower bounds. Using these bounds together with the preset ratio parameter $\alpha$, the Threshold Derivation module applies the LATS logic as defined in Eq. \eqref{eq:DFD} to compute the threshold $\eta_i$. Once computed, the derived threshold is broadcast to all bit-level PE lanes.

\begin{table}[t]
\renewcommand{\arraystretch}{0.93}
\centering
\scriptsize
\caption{Hardware Configurations of BitStopper}\vspace{-2.5mm}
\begin{tabular}{p{2.0cm}|m{5.7cm}<{\centering}}
\specialrule{0.12em}{0.5pt}{0.9pt}
\multirow{2}{*}{\textbf{Main Memory}} & HBM2; 8 channels $\times$ 128-bit at 2Gbps; \\
                    & each channel provides 32GB/s bandwidth. \\ \specialrule{0.05em}{0.5pt}{0.9pt}
\multirow{2}{*}{\textbf{On-chip Buffer}} & 320KB SRAM for Key and Value buffers; \\
                        & 8KB Q buffer. \\ \specialrule{0.05em}{0.5pt}{0.9pt}
\multirow{2}{*}{\textbf{QK-PU}} & 32 Bit-level PE Lanes\\
& An LATS module; A Bit Margin Generator\\
\specialrule{0.05em}{0.5pt}{0.9pt}
\multirow{2}{*}{\hspace{1mm}\textbf{- Bit-level PE Lane}} & \!\!64-dim\,$\times$\,12-bit\,$\times$\,1-bit Bit-serial Resuable ANDer tree; \\
                 & 64 entry $\times$ 45-bit Scoreboard. \\ \specialrule{0.05em}{0.5pt}{0.9pt}
\multirow{2}{*}{\textbf{V-PU}} & Single 1-D 64 way 12bit $\times$12bit MAC array; \\
                 & A 18-bit input, 18-bit output LUT-based Softmax.\\
\specialrule{0.12em}{0.5pt}{0.5pt}
\end{tabular}
\label{tab:BitStopper}\vspace{-5mm}
\end{table}

\section{Evaluation}\label{sec:Evaluation}
\subsection{Experimental Setup}
\textbf{Algorithm Evaluation}. We evaluate our pruning strategy on two representative LLMs: OPT1B3 \cite{zhang2022opt} and Llama2-7B \cite{touvron2023llama2}. Model quality is assessed via perplexity (PPL) on the Wikitext-2 dataset \cite{merity2016pointer} and Dolly \cite{conover2023free}, where lower PPL indicates better performance. All evaluations are conducted using Pytorch and the HuggingFace library \cite{wolf2020transformers}. To accommodate outlier values, we employ INT12 quantization, which serves as the accuracy baseline and is derived via post-training quantization (PTQ).

\textbf{Baseline and prior-art accelerators.} We compare BitStopper with three DS accelerators for attention: 1) Baseline: a dense version based on BitStopper without sparsity modules. 2) Sanger \cite{lu2021sanger}: a coarse-grained accelerator with 4-bit sparsity predictor. 3) SOFA \cite{wang2024sofa}: which uses a top-$k$-based predictor and employs cross-stage tiling to minimize DRAM access. However, it relies on costly retraining to recover accuracy. 4) TokenPicker\cite{park2024token}: which utilizes 4-bit chunk progressive pruning with post-exp probability.  For fairness, all designs are normalized to a 28nm process and evaluated under identical conditions: PE arrays occupy the same area as BitStopper and work in 1GHz, on-chip SRAM is set to 328kB.

\textbf{Hardware Evaluation}. Table \ref{tab:BitStopper} depicts the hardware configuration of the BitStopper accelerator. Each QK-PU comprises 32 PE lanes, with each lane processing 64 bits of a Key vector per cycle to fully utilize HBM2 bandwidth. BitStopper is implemented in RTL and synthesized using Synopsys DC under 28nm CMOS to estimate logic area and power (Fig. \ref{fig:Area_Power}). The speedup is evaluated via an in-house developed cycle-level simulator. The energy and area of on-chip buffers and scoreboard are estimated through CACTI \cite{balasubramonian2017cacti}. To get the number of cycles and energy of off-chip accesses, we use Ramulator \cite{luo2023ramulator} with trace files generated in RTL simulation.

\begin{figure}[t]
\centering
\includegraphics[width=\linewidth]{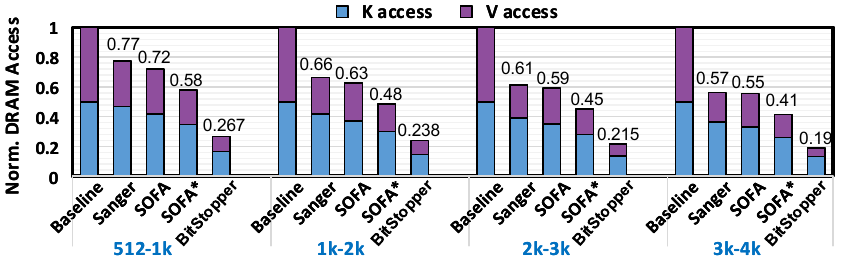}\vspace{-4mm}
\caption{Normalized off-chip memory access comparison for Llama-7B model. SOFA$^*$ performs additional fine-tuning on Dolly and Wikitext datasets.}
\label{fig:Lateral_memory_access_reduction}\vspace{-5mm}
\end{figure}

\begin{figure}[t]
\centering
\includegraphics[width=\linewidth]{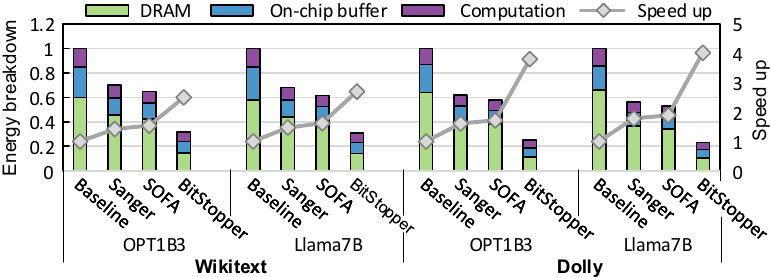}\vspace{-4mm}
\caption{Speedup and energy breakdown in (a) Wikitext-2 and (b) Dolly tasks.}
\label{fig:energy_breakdown}\vspace{-4mm}
\end{figure}

\begin{figure}[t]
\centering
\includegraphics[width=\linewidth]{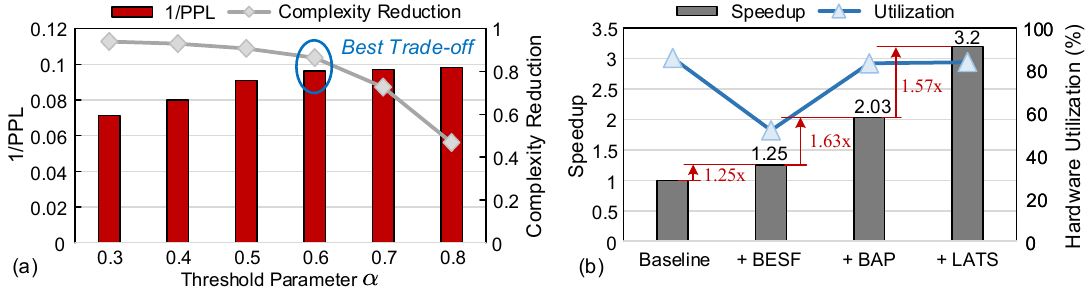}\vspace{-4mm}
\caption{1/PPL (higher is better) and complexity reduction versus pruning parameter $\alpha$. The results are derived from Llama7B on the Dolly dataset. (b) Speedup breakdown and hardware utilization.}
\label{fig:Speedup}\vspace{-7mm}
\end{figure}

\textbf{Configurations}. We evaluate the PPL with varying pruning ratio by tuning the parameter $\alpha$ from $0.2$ to $0.8$ in step of $0.1$. We then apply the successive halving method to expedite the process.  For all hardware evaluations, sequence lengths are set to $1k$ for OPT1.3B and $2k$ for LLaMA2-7B for the Wikitext dataset, while they are set to $2k$ and $4k$ for Dolly.

\subsection{Complexity Reduction Comparisons} 
Fig. \ref{fig:memory_access_reduction} compares the normalized complexity of different DS works under comparable PPL. Sanger predicts sparsity using an extra 4-bit predictor with static thresholding, whereas SOFA relies on a log-domain predictor combined with top-$k$ sorting. As shown, while Sanger and SOFA successfully alleviate computation complexity by an average of $69\%$ and $65\%$ (blue part) through token sparsity, respectively, they fail to realize evident memory access reduction (green part). Such IO burden becomes more prohibitive in long sequence scenarios (see Dolly task). This is because their sparsity predictions have to fetch full-size ($S$$\times$$H$) Key matrix, resulting in a heavy IO burden. While TokenPicker removes the extra predictor, its post-exp decision incurs significant computational overhead. Further, its coarse 4-bit chunk granularity misses finer-grained redundancy, leaving substantial optimization potential untapped. By contrast, BitStopper’s BESF and LATS strategies substantially reduce both IO and computation complexity while preserving comparable PPL. These gains stem from fine-grained, stage-fusion bit early stopping and adaptive decision-making, which eliminate more invalid processing.

Fig.~\ref{fig:Lateral_memory_access_reduction} compares DRAM access reduction across different designs and sequence lengths. For fairness, we set the precision of QKV to 12 bits and allows almost +$0.1$ PPL on Wikitext-2 and Dolly tasks. BitStopper consistently outperforms Sanger and SOFA*, achieving average reductions of 2.9× and 2.1×. This benefit stems from the synergy between the LATS and BAP strategies, which dynamically adjust thresholds and eliminate redundant memory accesses at fine granularity. In contrast, Sanger's coarse-grained filtering prevents early termination of trivial Key accesses, while SOFA's fixed top-$k$ selection lacks adaptability, degrading inference accuracy without fine-tuning. Notably, BitStopper achieves a 2.8× greater memory access reduction than unfinetuned SOFA.

\subsection{Performance Evaluation}
Fig. \ref{fig:energy_breakdown} compares the throughput and energy consumption. The energy is broken down into computation, on-chip buffer, and off-chip memory. Sanger ($67\%$) and SOFA ($62\%$) exhibit higher DRAM overheads due to additional prediction units, leading to substantial unreusable memory access. In contrast, BitStopper limits this overhead to $38\%$ via bit-grained stage-fusion strategy. In terms of speedup, as the sequence length increases, BitStopper's speedup becomes more pronounced. This is because longer sequences contain more redundancy, and the distribution disparity among different queries increases. BitStopper's LATS strategy is well-suited to handle such scenarios. In contrast, static strategies in Sanger and SOFA result in redundant computation and memory access. Overall, BitStopper achieves average speedups of $3.2\times$, $2.03\times$, and $1.89\times$, and energy efficiency improvements of $3.7\times$, $2.4\times$ and $2.1\times$, compared to Baseline, Sanger, and SOFA, respectively. 


Fig. \ref{fig:Speedup} (a) analyzes how different pruning parameters $\alpha$ affect PPL and reduced complexity. Overall, a smaller $\alpha$ delivers more pruning, yielding greater complexity reduction but lowering 1/PPL. As observed, when $\alpha$ falls below $0.6$, the reduction in complexity begins to plateau, while 1/PPL drops sharply. This is because overly aggressive token pruning, which removes certain critical tokens, thus severely degrading the 1/PPL and hindering further complexity reduction. To strike a balanced trade-off between accuracy and efficiency, we thus prioritize setting the value of $\alpha$ near $0.6$. 

Fig. \ref{fig:Speedup} (b) presents a breakdown of the speedup achieved by BitStopper, using a baseline dense attention accelerator derived from BitStopper but excluding all sparse processing modules. While the incorporation of BESF yields a modest throughput improvement, compute unit utilization remains limited to $48\%$ due to unhidden memory access latency, thereby capping the overall speedup at $1.25\times$. With the integration of BAP, asynchronous out-of-order execution effectively mitigates memory latency, increasing compute unit utilization to $83\%$ and enhancing the speedup by $1.63\times$. Finally, LATS reduces memory accesses and computations for irrelevant tokens, lowering dynamic power consumption and achieving a $1.57\times$ speedup.

\subsection{Area and Power}
\begin{figure}[t]
\centering
\includegraphics[width=\linewidth]{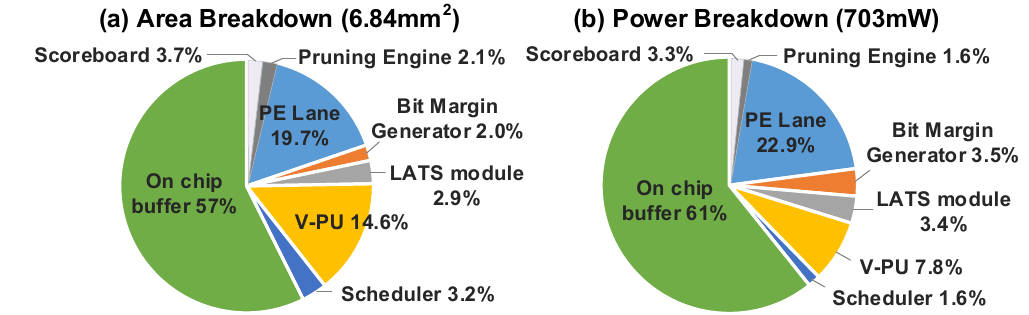}\vspace{-2mm}
\caption{Area/Power of BitStopper at TSMC 28nm, 1GHz.}
\label{fig:Area_Power}\vspace{-4mm}
\end{figure}

Fig. \ref{fig:Area_Power} presents the area and power breakdown of the BitStopper accelerator. Occupying 6.84 mm$^2$ and consuming 703 mW, it achieves a peak energy efficiency of 11.36 TOPS/W. The newly added Bit Margin Generator and LATS modules adaptively respond to attention distribution and enhance token pruning, incurring only $4.9\%$ area and $6.9\%$ power overhead. Additionally, the integration of the Scoreboard and Pruning Engine into PE lanes enables stage fusion, adding $5.8\%$ area and $4.9\%$ power overhead. Despite the modest hardware cost, the elimination of the sparsity predictor and reduced off-chip memory access lead to significant efficiency gains. BitStopper thus represents a deliberate trade-off, achieving substantial energy efficiency improvements with minimal resource overhead. 

\section{Related Works}\label{sec:Related Works}
Numerous attention accelerators \cite{li2020ftrans, ham20203, ham2021elsa, qu2022dota,  yang2022dtatrans, hong2022dfx, liu2024hsconn, wang2024sofa,  li2022accelerating, lu2021sanger, wang2021spatten, zhou2022energon, qin2023fact, fan2022adaptable, zhao2024hardware, park2024token,wang2025beta} have been proposed. Early works, such as A$^3$ \cite{ham20203} and ELSA \cite{ham2021elsa} accelerate computation via approximation techniques. Recent efforts have shifted to jointly optimizing computation and memory. Energon \cite{zhou2022energon} and SOFA \cite{wang2024sofa} adopt fine-grained filtering and tiling to alleviate memory overhead. However, they still rely on extra sparsity predictors, which become de facto latency and power bottlenecks after sparsification. Although the more recent BETA \cite{wang2025beta} introduces bit-level filtering, it likewise retains an additional sparsity predictor. Further, the absence of a bit-uncertainty margin renders its pruning decisions highly inaccurate. BitStopper is the first to explicitly identify and address this issue. It removes the need for external predictors by fine-grained, bit-serial computation, while leveraging efficient bit-level early termination and reuse.

Although TokenPicker \cite{park2024token} also adopts bit-chunk reuse to alleviate memory access, BitStopper differs in several key aspects: 1) Simpler decision strategy. BitStopper employs a lightweight, max-based decision, whereas TokenPicker relies on a more complex Softmax-based selection. 2) Broader applicability. Its simplified decision logic allows BitStopper to operate in both Prefill and Decoding, whereas TokenPicker is restricted to Decoding only. 3) Finer granularity. BitStopper operates at a bit-level reuse granularity, while TokenPicker is constrained to 4-bit chunks, making it conceptually closer to previous works such as SpAtten \cite{wang2021spatten} and Energon \cite{zhou2022energon}.  

\section{Conclusion}\label{sec:conclusion}
This paper presents BitStopper, a bit-grained, stage-fusion accelerator that eliminates the extra sparsity predictor via bit-level reuse opportunities. A lightweight adaptive token selection strategy improves pruning accuracy, while bit-level asynchronous processing enhances hardware utilization. Experimental results show that BitStopper delivers superior efficiency compared with existing attention sparsity accelerators.

\section*{Acknowledgments}
This work was supported in part by the National Science and Technology Major Project under Grant 2022ZD0115200; in part by the NSFC under Grant 62125403, Grant 92464302, Grant U24B20164 and Grant 92164301; in part by Shanghai Municipal Science and Technology Major Project; in part by the Natural Science Foundation of Jiangsu Province Basic Research Program under Grant BK20243042; in part by the Beijing National Research Center for Information Science and Technology; in part by the Northern IC Technology Innovation Center (Beijing) Co., Ltd under Grant QYJS20232801B; and in part by the Beijing Advanced Innovation Center for Integrated Circuits.

\bibliographystyle{IEEEtran}
\footnotesize
\bibliography{IEEEabrv,main}

\end{document}